\documentclass[10pt,twocolumn,letterpaper]{article}

\usepackage{iccv}
\usepackage{times}
\usepackage{epsfig}
\usepackage{graphicx}
\usepackage{amsmath}
\usepackage{amssymb}

\usepackage{microtype}
\usepackage{arydshln}
\usepackage{amsthm}
\usepackage{multirow}
\usepackage{braket}
\usepackage{algorithm}
\usepackage{algorithmic}

\def\Eq#1{Eq.(\ref{eqn:#1})}
\def\Fig#1{Fig.\ref{fig:#1}}
\def\Alg#1{Algorithm~\ref{alg:#1}}
\def\Table#1{Table~\ref{table:#1}}
\def\Sec#1{Section~\ref{sec:#1}}

\usepackage[pagebackref=true,breaklinks=true,letterpaper=true,colorlinks,bookmarks=false]{hyperref}

\iccvfinalcopy 

\def\iccvPaperID{1230} 

\ificcvfinal\fi
\begin{document}

\ificcvfinal
\renewcommand{\baselinestretch}{0.96}
\setlength\textfloatsep{13pt}
\setlength\floatsep{5pt}
\setlength\abovecaptionskip{1truemm}
\else
\fi

\title{Temporal Generative Adversarial Nets with Singular Value Clipping}

\author{Masaki Saito\thanks{Authors contributed equally}
~~~~~~~~ Eiichi Matsumoto$^*$
~~~~~~ Shunta Saito\\
Preferred Networks inc., Japan\\
{\tt\small \{msaito, matsumoto, shunta\}@preferred.jp}
}

\maketitle

\begin{abstract}
In this paper, we propose a generative model,
Temporal Generative Adversarial Nets (TGAN),
which can learn a semantic representation of unlabeled videos,
and is capable of generating videos.
Unlike existing Generative Adversarial Nets (GAN)-based methods
that generate videos with a single generator consisting of 3D deconvolutional layers,
our model exploits two different types of generators:
a temporal generator and an image generator.
The temporal generator takes a single latent variable
as input and outputs a set of latent variables,
each of which corresponds to an image frame in a video.
The image generator transforms a set of such latent variables into a video.
To deal with instability in training of GAN with such advanced networks,
we adopt a recently proposed model, Wasserstein GAN,
and propose a novel method to train it stably in an end-to-end manner.
The experimental results demonstrate the effectiveness of our methods.
\end{abstract}

\section{Introduction}
Unsupervised learning of feature representation from a large dataset is
one of the most significant problems in computer vision.
If good representation of data can be obtained from an unlabeled dataset,
it could be of benefit to a variety of tasks such as
classification, clustering, and generating new data points.

There have been many studies regarding unsupervised learning in the field of computer vision.
Their targets are roughly two-fold; images and videos.
As for unsupervised learning of images,
Generative Adversarial Nets (GAN) \cite{Goodfellow14}
have shown impressive results and
succeeded to generate plausible images with a dataset that
contains plenty of natural images \cite{Denton15,Wang16}.
In contrast, unsupervised learning of videos still has many difficulties compared to images.
While recent studies have achieved remarkable progress
\cite{Srivastava15,Oh15,Kalchbrenner16}
in a problem that predicts future frames from previous frames,
video generation without any clues of data is still a highly challenging problem.
Although the recent study tackled to address this problem
by decomposing it into background generation and foreground generation,
this approach has a drawback that
it cannot generate a scene with dynamic background
due to the static background assumption \cite{Vondrick16}.
To the best of our knowledge,
there is no study that tackles video generation without such assumption and
generates diversified videos like natural videos.

Although a simple approach is to use 3D convolutional layers
for representing the generating process of a video,
it implies that images along x-t plane and y-t plane besides x-y plane are considered equally,
where x and y denote the spatial dimensions and t denotes the time dimension.
We believe that the nature of time dimension is essentially different
from the spatial dimensions in the case of videos
so that such approach has difficulty on the video generation problem.
The relevance of this assumption has been also discussed in some recent studies \cite{Simonyan14,Ng15,Wang15a}
that have shown good performance on the video recognition task.

\begin{figure*}[t]
  \centering
  \includegraphics[width=1.0\linewidth]{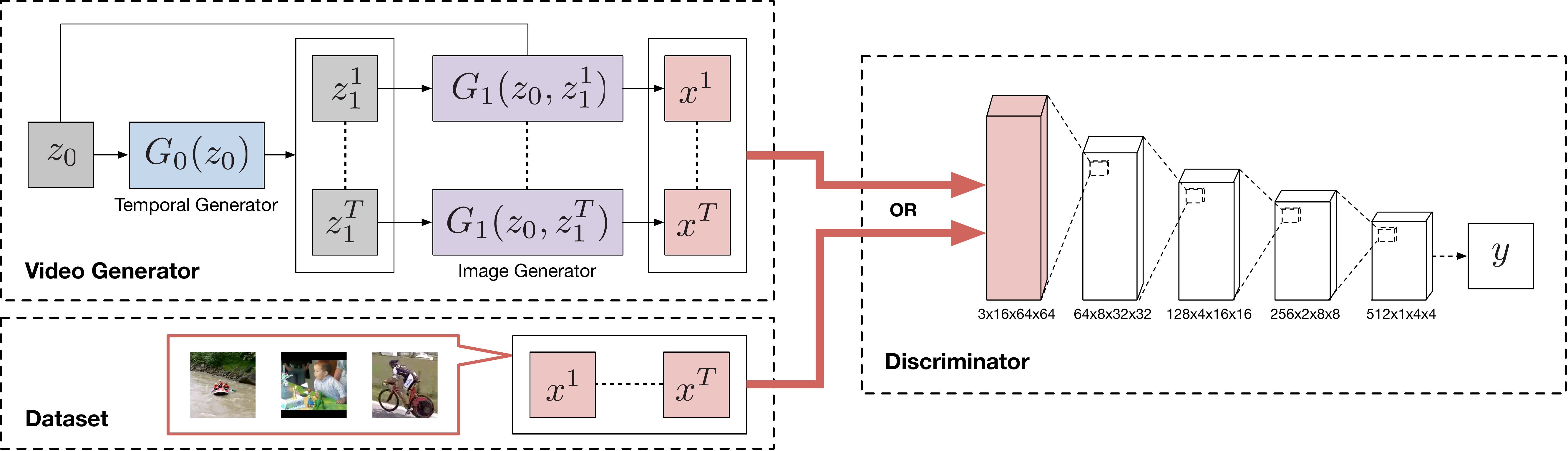}
  \caption{Illustration of TGAN.
  The video generator consists of two generators,
  the temporal generator and the image generator.
  The temporal generator $G_0$ yields a set of latent variables $z_1^t (t=1,\dots,T)$ from $z_0$.
  The image generator $G_1$ transforms those latent variables $z_1^t (t=1,\dots,T)$ and $z_0$ into a video data which has $T$ frames.
  The discriminator consists of three-dimensional convolutional layers, and
  evaluates whether these frames are from the dataset or the video generator. The shape of a tensor in the discriminator is
  denoted as ``(channels)$\times$(time)$\times$(height)$\times$(width)''.
  }
  \label{fig:diagram}
  \vspace{-7pt}
\end{figure*}
Based on the above discussion, in this paper, we extend an existing GAN
model and propose Temporal Generative Adversarial Net (TGAN)
that is capable of learning representation
from an unlabeled video dataset and producing a new video.
Unlike the existing video generator that generates videos with 3D deconvolutional
layers \cite{Vondrick16}, in our proposed model the generator consists of two
sub networks called a {\it temporal generator} and an {\it image generator}
(\Fig{diagram}). Specifically, the temporal generator first yields a set of
latent variables, each of which corresponds to a latent variable for the image
generator. Then, the image generator transforms these latent variables into
a video which has the same number of frames as the variables.
The model comprised of the temporal and image generators can not only enable to 
efficiently capture the time series,
but also be easily extended to frame interpolation.
The typical problem that arises from such advanced networks is the instability of training of GANs.
%
In this paper we adopt a recently proposed Wasserstein GAN (WGAN) which tackles the instability, however, we observed that our model still has sensitivity
to a hyperparameter of WGAN.
%
Therefore, to deal with this problem,
we propose a novel method to remove the sensitive hyperparameter from WGAN
and stabilize the training further.
The experiments show that our method is more stable than the conventional methods,
and the model can be successfully trained even under the situation
where the loss diverges with the conventional methods.

Our contributions are summarized as follows.
(i) The generative model that can efficiently capture
the latent space of the time dimension in videos.
It also enables a natural extension to an application such as frame interpolation.
(ii) The alternative parameter clipping method for WGAN that significantly stabilizes
the training of the networks that have advanced structure.
%
%
\section{Related work}
\label{sec:Related work}

\subsection{Natural image generation}
Supervised learning with Convolutional Neural Networks (CNNs)
has recently shown outstanding performance
in many tasks such as image classification \cite{He15,He16,Huang16}
and action recognition \cite{Ji13,Karpathy14,Simonyan14,Tran15},
whereas unsupervised learning with CNN has received relatively less attention.
A common approach for generating images is the use of undirected
graphical models such as Boltzmann machines
\cite{Salakhutdinov09,Lee09a,Eslami12}.
However, due to the difficulty of approximating gradients,
it has been empirically observed that
such deep graphical models frequently fail to find
good representation of natural images with sufficient diversity.
Both Gregor \etal \cite{Gregor15} and Dosvotiskiy \etal \cite{Dosovitskiy14}
have proposed models that respectively use recurrent and
deconvolutional networks, and successfully generated natural images.
However, both models make use of supervised learning and require
additional information such as labels.

The Generative Adversarial Network (GAN),
which we have mainly employed in this study,
is a model for unsupervised learning that
finds a good representation of samples
by simultaneously training two different networks called
the {\it generator} and the {\it discriminator}.
Recently, many extensions for GANs have been proposed.
Conditional GANs performs modeling of object attributes \cite{Mirza14,Im16}.
Pathak \etal \cite{Pathak16} adopted the adversarial network
to generate the contents of an image region conditioned on its surroundings.
Li and Wand \cite{Li16} employed the GAN model
in order to efficiently synthesize texture.
Denton \etal \cite{Denton15} proposed a Laplacian GAN that outputs a
high-resolution image by iteratively generating images
in a coarse-to-fine manner.
Arjovsky \etal \cite{Arjovsky17} transformed the training of GAN into
the minimization problem of Earth Mover's distance, and proposed 
a more robust method to train both the generator and the discriminator.
Radford \etal \cite{Radford16} also proposed a simple yet powerful model
called Deep Convolutional GAN (DCGAN) for generating realistic images
with a pair of convolutional and deconvolutional networks.
Based on these results, Wang \etal \cite{Wang16} extended
DCGAN by factorizing the image generating process into two paths,
and proposed a new model called a Style and Structure GAN ($\mathrm{S}^2$-GAN)
that exploits two types of generators.

\subsection{Video recognition and unsupervised learning}
As recognizing videos is a challenging task which has received a lot of
attention, many researchers have tackled this problem in various ways.
In supervised learning of videos, while a common approach is to use
dense trajectories \cite{Wang11a,Sadanand12,Rohrbach12},
recent methods have employed CNN and achieved state-of-the-art results
\cite{Ji13,Karpathy14,Simonyan14,Tran15,Ng15,Wang15a,Wang16a}.
Some studies are focused on extracting spatio-temporal feature vectors
from a video in an unsupervised manner.
Taylor \etal \cite{Taylor10} proposed a method that extracts invariant features
with Restricted Boltzmann Machines (RBMs).
Temporal RBMs have also been proposed to explicitly capture
the temporal correlations in videos \cite{Taylor07,Sutskever07,Sutskever08}.
Stavens and Thrun \cite{Stavens10} dealt with this problem
by using an optical flow and low-level features such as SIFT.
Le \etal \cite{Le11a} use Independent Subspace Analysis (ISA)
to extract spatio-temporal semantic features.
Deep neural networks have also been applied to feature extraction from videos
\cite{Zou12,Goroshin15,Wang15} in the same way as supervised learning.

There also exist several studies focusing on predicting video sequences
from an input sequence with Recurrent Neural Networks (RNNs)
represented by Long Short-Term Memory (LSTM) \cite{Hochreiter97}.
In particular, Ranzato \etal \cite{Ranzato14} proposed a
Recurrent Neural Network (RNN) model that can learn both
spatial and temporal correlations.
Srivastava \etal \cite{Srivastava15} also applied LSTMs and
succeeded to predict the future sequence of a simple video.
Zhou and Berg \cite{Zhou16} proposed a network that creates
depictions of objects at future times with LSTMs and DCGAN.
Kalchbrenner \etal \cite{Kalchbrenner16} also employed
a convolutional LSTM model, and proposed Video Pixel Networks that
directly learn the joint distribution of the raw pixel values.
Oh \etal \cite{Oh15} proposed a deep auto-encoder model conditioned on actions,
and predicted next sequences of Atari games from a single screen shot and
an action sent by a game pad.
In order to deal with the problem that generated sequences are
``blurry'' compared to natural images, 
Mithieu \etal \cite{Mithieu16} replaced a standard mean squared error loss
and improved the quality of predicted images.
However, the above studies cannot directly be applied to the task of generating
entire sequences from scratch
since they require an initial sequence as an input.

Vondrick \etal \cite{Vondrick16} recently proposed a generative model 
that yields a video sequence from scratch with DCGAN
consisting of 3D deconvolutional layers.
The main difference between their model and ours is {\it model representation};
while they simplified the video generation problem by assuming that
a background in a video sequence is always static and
generate the video with 3D deconvolutions,
we do not use such assumption and decompose the generating process of video into
the 1D and 2D deconvolutions.

\section{Temporal Generative Adversarial Nets}
%
\subsection{Generative Adversarial Nets}
Before we go into the details of TGAN, we briefly explain the existing GAN
\cite{Goodfellow14} and the Wasserstein GAN \cite{Arjovsky17}.
A GAN exploits two networks called the generator and the discriminator.
The generator $G: \mathbb{R}^{K} \to \mathbb{R}^{M}$
is a function that generates samples $x \in \mathbb{R}^{M}$
which looks similar to a sample in the given dataset.
The input is a latent variable
$z \in \mathbb{R}^{K}$, where
$z$ is randomly drawn from a given distribution $p_{G}(z)$, e.g.,
a uniform distribution.  The discriminator
$D: \mathbb{R}^{M} \to [0, 1]$ is a classifier that discriminates
whether a given sample is from the dataset or generated by $G$.

The GAN simultaneously trains the two networks by playing a non-cooperative
game; the generator wins if it generates an image that the discriminator
misclassifies, whereas the discriminator wins if it correctly classifies
the input samples.
Such minimax game can be represented as
\begin{multline}
\label{eqn:GAN}
\min_{\theta_{G}} \max_{\theta_{D}} \;
\mathbb{E}_{x \sim p_\mathrm{data}}[\ln D(x)] \\
+ \mathbb{E}_{z \sim p_{G}} [\ln(1 - D(G(z)))],
\end{multline}
where $\theta_{G}$ and $\theta_{D}$ are the parameters of the generator
and the discriminator, respectively.
$p_{\mathrm{data}}$ denotes the empirical data distribution.



\subsection{Wasserstein GAN}
It is known that the GAN training is unstable and requires careful adjustment
of the parameters. To overcome such instability of learning,
Arjovsky \etal \cite{Arjovsky17} focused on the property that the GAN training
can also be interpreted as the minimization of the Jensen-Shannon (JS) divergence,
and proposed Wasserstein GAN (WGAN) that trains the generator and the discriminator
to minimize an Earth Mover's distance
(EMD, a.k.a. first Wasserstein distance) instead of the JS divergence.
Several experiments the authors conducted reported that
WGANs are more robust than ordinal GANs, and tend to avoid mode dropping.

The significant property in the learning of WGAN is
``$K$-Lipschitz'' constraint with regard to the discriminator.
Specifically, if the discriminator satisfies the $K$-Lipschitz constraint,
i.e., 
$\vert D(x_1) - D(x_2) \vert \leq K \vert x_1 - x_2 \vert$
for all $x_1$ and $x_2$,
the minimax game of WGAN can be represented as
\begin{align}
\label{eqn:WGAN}
\min_{\theta_{G}} \max_{\theta_{D}}
\mathbb{E}_{x \sim p_\mathrm{data}}[D(x)]
- \mathbb{E}_{z \sim p_{G}} [D(G(z))].
\end{align}
Note that unlike the original GAN,
the return value of $D$ in \Eq{WGAN} is an unbounded real value, i.e., $D: \mathbb{R}^{M} \to \mathbb{R}$.
In this study we use \Eq{WGAN} for training instead of \Eq{GAN}.

In order to make the discriminator be the $K$-Lipschitz,
the authors proposed a method that clamps all the weights
in the discriminator to a fixed box denoted as $w \in [-c, c]$.
Although this weight clipping is a simple and assures the discriminator satisfies the $K$-Lipschitz condition,
it also implies we cannot know the relation of the parameters
between $c$ and $K$. As it is known that the objective of the 
discriminator of \Eq{WGAN} is a good approximate expression of EMD in the case of $K = 1$, this could be a problem when we want to find the approximate value of EMD.



\subsection{Temporal GAN}
Here we introduce the proposed model based on the above discussion.
Let $T > 0$ be the number of frames to be generated, and
$G_0 : \mathbb{R}^{K_0} \to \mathbb{R}^{T \times K_1}$
be the temporal generator that gets another latent variable
$z_0 \in \mathbb{R}^{K_0}$ as an argument and generates latent variables
denoted as $[z_1^1, \dots, z_1^T]$.
In our model, $z_0$ is randomly drawn from a distribution $p_{G_0}(z_0)$.

Next, we introduce {\it image generator}
$G_1 : \mathbb{R}^{K_0} \times \mathbb{R}^{K_1} \to \mathbb{R}^{M}$
that yields a video from these latent variables. Note that $G_1$
takes both the latent variables generated from $G_0$ as well as
original latent variable $z_0$ as arguments.
While $z_1$ varies with time, $z_0$ is invariable regardless
of the time, and we empirically observed 
that it has a significant role in suppressing a sudden change of the action
of the generated video. That is, in our representation, the generated video is
represented as $[G_1(z_0, z_1^1), \dots, G_1(z_0, z_1^T)]$.


Using these notations, \Eq{WGAN} can be rewritten as
\begin{multline}
\label{eqn:TemporalGAN}
\min_{\theta_{G_0},\theta_{G_1}} \max_{\theta_{D}} \;\;
\mathbb{E}_{[x^1, \dots, x^T] \sim p_\mathrm{data}}
[D([x^1, \dots, x^T ])] \\
- \mathbb{E}_{z_0 \sim p_{G_0}} [D([G_1(z_0, z_1^1), \dots, G_1(z_0, z_1^T) ])] \big),
\end{multline}
where $x^t$ is the $t$-th frame of a video in a dataset,
and $z_1^t$ is the latent variable corresponding to $t$-th frame generated by $G_0(z_0)$.
$\theta_{D}$, $\theta_{G_0}$, and $\theta_{G_1}$ represent the parameter of
$D$, $G_0$, and $G_1$, respectively.

\subsection{Network configuration}
\begin{table}[t]
{\renewcommand{\arraystretch}{1.1}
\centering
\begin{tabular}{|c|c|c|}
\hline
Temporal generator & \multicolumn{2}{|c|}{Image generator} \\ \hline \hline
$z_0 \in \mathbb{R}^{1 \times 100}$ &
$z_0 \in \mathbb{R}^{1 \times 100}$ & $z_1^t \in \mathbb{R}^{100}$ \\ \hline
deconv (1, 512, 0, 1) & {\small linear ($256 \cdot 4^2$)} & {\small linear ($256 \cdot 4^2$)} \\ \cline{2-3}
deconv (4, 256, 1, 2) & \multicolumn{2}{|c|}{concat + deconv (4, 256, 1, 2)} \\
deconv (4, 128, 1, 2) & \multicolumn{2}{|c|}{deconv (4, 128, 1, 2)} \\
deconv (4, 128, 1, 2) & \multicolumn{2}{|c|}{deconv (4, 64, 1, 2)} \\
deconv (4, 100, 1, 2) & \multicolumn{2}{|c|}{deconv (4, 32, 1, 2)} \\
tanh & \multicolumn{2}{|c|}{deconv (3, 3, 1, 1) + tanh} \\ \hline
\end{tabular}
}
\vspace{2pt}
\caption{Network configuration of the generator.
The second row represents the input variables.
``linear ($\cdot$)'' is the number of output units in the linear layer.
The parameters in the convolutional and the deconvolutional layer are denoted as
``conv/deconv ((kernel size), (output channels), (padding), (strides)).''}
\label{table:TemporalNets}
\end{table}
\label{sec:Configuration}
This subsection describes the configuration of our three networks:
the temporal generator, the image generator, and the discriminator.
\Table{TemporalNets} shows a typical network setting.


\paragraph{Temporal generator}
Unlike typical CNNs that perform two-dimensional convolutions in the spatial direction,
the deconvolutional layers in the temporal generator perform a one-dimensional deconvolution in the temporal direction.
For convenience of computation, we first regard $z_0 \in \mathbb{R}^{K_0}$ as a one-dimensional activation map of 
$z_0 \in \mathbb{R}^{1 \times K_0}$, where the length and the number of channels are one and $K_0$, respectively.
A uniform distribution is used to sample $z_0$.
Next, applying the deconvolutional layers we expand its length while reducing the number of channels.
The settings for the deconvolutional layers are the same as those of the image generator except for the number of channels and one-dimensional deconvolution.
Like the original image generator we insert a Batch Normalization (BN) layer \cite{Ioffe15}
after deconvolution and use Rectified Linear Units (ReLU) \cite{Nair10} as activation functions.


\paragraph{Image generator}
The image generator takes two latent variables
as arguments. After performing a linear transformation on each variable, we reshape them into the form shown in \Table{TemporalNets}, concatenate them and perform five deconvolutions.
These settings are almost the same as the existing DCGAN,
i.e., we used ReLU \cite{Nair10} and Batch Normalization layer \cite{Ioffe15}.
The kernel size, stride, and padding are respectively 4, 2, and 2
except for the last deconvolutional layer.
Note that the number of output channels of the last deconvolutional layer
depends on whether the dataset contains color information or not.

\paragraph{Discriminator}
We employ spatio-temporal 3D convolutional layers to model the discriminator.
The layer settings are similar to the image generator.
Specifically, we use four convolutional layers with
$4 \times 4 \times 4$ kernel and a stride of 2.
The number of output channels is 64 in the initial convolutional layer,
and set to double when the layer goes deeper.
As with the DCGAN, we used LeakyReLU \cite{Maas13} with $a = 0.2$
and Batch Normalization layer \cite{Ioffe15} after these convolutions.
Note that we do not insert the batch normalization after the initial convolution.
Finally, we use a fully-connected layer and summarize
all of the units in a single scalar.
Each shape of the tensor used in the discriminator is shown in \Fig{diagram}.

\section{Singular Value Clipping}
As we described before, WGAN requires the discriminator to fulfill
the $K$-Lipschitz constraint, and the authors employed a parameter
clipping method that clamps the weights in the discriminator to $[-c, c]$.
However, we empirically observed that the tuning of hyper parameter $c$ is
severe, and it frequently fails in learning under a different
situation like our proposed model. 
We assumed this problem would be caused by a property that the $K$-Lipschitz 
constraint widely varies depending the value of $c$,
and propose an alternative method that can explicitly adjust the value of $K$.

Suppose that $D(x)$ is a composite function consisting of $N$ primitive functions, and each function $f_n$ is Lipschitz continuous with $K_n$.
In this case $D$ can be represented as
$D = f_N \circ f_{N-1} \circ \cdots f_1$, and $D$ is also Lipschitz continuous
with $K = \prod_n K_n$. That is, what is important in our approach is to 
add constraints to all the functions such that $f_n$ satisfies the condition of 
given $K_n$.
Although in principle our method can derive operations that satisfy arbitrary 
$K$, in the case of $K=1$ these operations are invariant regardless of
the number of layers constituting the discriminator.
For simplicity we focus on the case of $K = 1$.

\begin{table}
\centering
{\renewcommand{\arraystretch}{1.1}
\begin{tabular}{|c|c|c|}
\hline
Layer & Condition & Method \\ \hline \hline
Linear & $\Vert W \Vert \leq 1$ & SVC \\
Convolution & $\Vert \hat{W} \Vert \leq 1$ & SVC \\
Batch normalization & $0 < \gamma \leq \sqrt{\sigma^2_B + \epsilon}$ & Clipping $\gamma$ \\
LeakyReLU & $a \leq 1$ & Do nothing \\ \hline
\end{tabular}
}
\vspace{0pt}
\caption{Proposed methods to satisfy the $1$-Lipschitz constraint.
$\Vert \cdot \Vert$ denotes a spectral norm. $a$ represents a fixed
parameter of the LeakyReLU layer. $\gamma$ and $\sigma_B$ are a scaling parameter after the batch normalization and a running mean of a standard deviation of a batch, respectively.}
%
%
\label{table:SingularValueClipping}
\end{table}

To satisfy $1$-Lipschitz constraint,
we add a constraint to all linear layers in the discriminator
that satisfies the spectral norm of weight parameter $W$ is equal
or less than one. This means that the singular values of weight matrix
are all one or less. To this end,
we perform singular value decomposition (SVD) after parameter update,
replace all the singular values larger than one with one,
and reconstruct the parameters with them.
We also apply the same operation to convolutional layers by
interpreting a higher order tensor in weight parameter as a matrix $\hat{W}$.
We call these operations {\it Singular Value Clipping} (SVC).

As with the linear and the convolutional layer,
we clamp the value of $\gamma$ which represents a scaling parameter
of the batch normalization layer in the same way.
We summarize a clipping method of each layer in \Table{SingularValueClipping}.
Note that we do not perform any operations on ReLU and LeakyReLU layers
because they always satisfy the condition unless $a$ in the LeakyReLU is
lower than 1.

\begin{figure}[t]
\centering
  \includegraphics[width=1.0\linewidth]{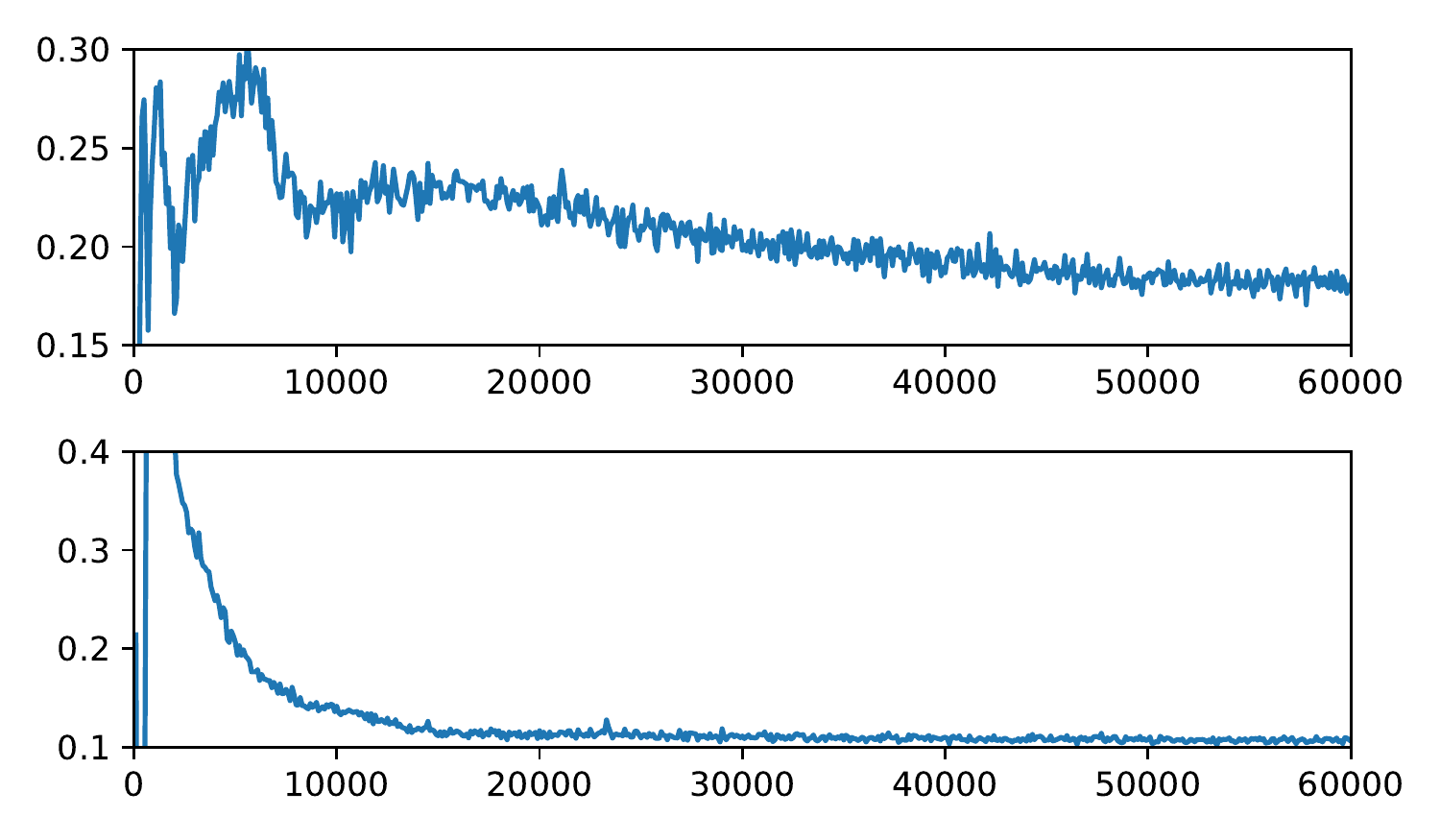}
  \caption{The difference of training curves in UCF-101 (see \Sec{Datasets} for details). The upper row shows the loss of the generator per iteration in conventional clipping method, while the lower row shows the loss in our clipping method, Singular Value Clipping.
  }
  \label{fig:SingularValueClipping}
\vspace{-3pt}
\end{figure}
\begin{algorithm}[t]
\caption{WGAN using Singular Value Clipping}
\label{alg:SingularValueClipping}
\begin{algorithmic}
\REQUIRE $\alpha$: the learning rate. $T$: the number of iterations. $n_D$: the number of iterations of the discriminator per generator's iteration. $n_{\mathrm{clip}}$: the number of intervals of the clipping.
%
%
\FOR{$t = 1$ to $T$}
\FOR{$n = 1$ to $n_{D}$}
\STATE Compute gradient of discriminator $g_D$
\STATE $\theta_D \gets \theta_D + \alpha \cdot \mathrm{RMSProp}(\theta_D, g_D)$
\ENDFOR
\STATE Compute gradient of generator $g_G$
\STATE $\theta_G \gets \theta_G - \alpha \cdot \mathrm{RMSProp}(\theta_G, g_G)$
\IF{$t \bmod n_{\mathrm{clip}} = 1$}
\STATE $\theta_D \gets \mathrm{SingularValueClipping}(\theta_D)$
\ENDIF
\ENDFOR
\end{algorithmic}
\end{algorithm}
The clear advantage of our alternative clipping method is that
it does not require the careful tuning of hyperparameter $c$.
Another advantage we have empirically observed is to stabilize the
training of WGAN; in our experiments,
our method can successfully train an advanced model even under the situation
where the behavior of loss function becomes unstable with the conventional clipping.
We show an example of such differences in \Fig{SingularValueClipping}.

Although the problem of SVC is an increased computational cost,
it can be mitigated by decreasing the frequency of performing the SVC.
We show the summary of the algorithm of WGAN with the SVC in
\Alg{SingularValueClipping}.
In our experiments, the computational time of SVD is almost the same as
that of the forward-backward computation,
but we observed the frequency of clipping is sufficient once every five iterations,
i.e., $n_{\mathrm{clip}} = 5$.

\section{Applications}
\label{sec:Applications}

\subsection{Frame interpolation}
\label{sec:FrameInterpolation}
One of the advantages of our model is to be able to generate
an intermediate frame between two adjacent frames.
Since the video generation in our model is formulated
as generating a trajectory in the latent image space represented by $z_0$ and $z_1^t$,
our generator can easily yield long sequences by just interpolating the trajectory.
Specifically, we add a bilinear filter to the last layer of the temporal generator, and interpolate the trajectory in the latent image space (see \Sec{Configuration}).

\subsection{Conditional TGAN}
In some cases, videos in a dataset contain some labels which
correspond to a category of the video such as ``IceDancing'' or ``Baseball''.
In order to exploit them and improve the quality of videos by the generator,
we also develop a Conditional TGAN (CTGAN),
in which the generator can take both label $l$ and latent variable $z_0$.

The structure of CTGAN is similar with that of the original Conditional
GAN. In temporal generator, after transforming $l$ into one-hot vector
$v_l$, we concatenate both this vector and $z_0$, and regard it as a new
latent variable. That is, the temporal generator of the CTGAN is denoted
as $G_0(z_0, v_l)$.
The image generator of the CTGAN also takes the one-hot label vector
as arguments, i.e., $G_1(z_0, z_1^t, v_l)$.
As with the original image generator, we first perform linear
transformation on each variable, reshape them, and operate five
deconvolutions.

In the discriminator, we first broadcast the one-hot label vector
to a voxel whose resolution is the same as that of the video.
Thus, if the number of elements of $v_l$ is $V$,
the number of channels of the voxel is equal to $V$.
Next, we concatenate both the voxel and the input video,
and send it into the convolutional layers.

\section{Experiments}

\subsection{Datasets}
\label{sec:Datasets}
We performed experiments with the following datasets.


\paragraph{Moving MNIST}
To investigate the properties of our models,
we trained the models on
the moving MNIST dataset \cite{Srivastava15}, in which there are 10,000 clips each of which has 20 frames and consists of
two digits moving inside a $64 \times 64$ patch.
In these clips, two digits move linearly and the direction and magnitude of motion vectors are randomly chosen.
If a digit approaches one of the edges in the patch,
it bounces off the edge and its direction is changed
while maintaining the speed. In our experiments,
we randomly extracted $16$ frames from these clips and used
them as a training dataset.

\paragraph{UCF-101}
UCF-101 is a commonly used video dataset that consists of 13,320 videos 
belonging to 101 different categories such as {\it IceDancing} and {\it Baseball Pitch} \cite{Soomro12}.
Since the resolution of videos in the dataset is too large
for the generative models, we resized all the videos to
$85 \times 64$ pixels, randomly extracted 16 frames,
and cropped a center square with 64 pixels.

\paragraph{Golf scene dataset}
Golf scene dataset is a large-scale video dataset made by Vondrick \etal \cite{Vondrick16}, and contains 20,268 golf videos with
$128 \times 128$ resolution.
Since each video includes 29 short clips on average,
it contains 583,508 short video clips in total.
As with the UCF-101, we resized all the video clips with $64 \times 64$ pixels.
To satisfy the assumption that the background is always fixed,
they stabilized all of the videos with SIFT and RANSAC algorithms.
As such assumption is not included in our method,
this dataset is considered to be advantageous for existing methods.

\subsection{Training configuration}
All the parameters used in the optimizer are the same as those of the
original WGAN. Specifically, we used the RMSProp optimizer
\cite{Tieleman12} with the learning rate of $0.00005$.
All the weights in the temporal generator and the discriminator are initialized with HeNormal \cite{He15}, and the weights in the image generator are initialized with the uniform distribution within a range of $[-0.01, 0.01]$.
Chainer \cite{Chainer15} was used to implement all models and for experiments.

For comparison, we employed the conventional clipping method and the SVC
to train models with the WGAN. In the conventional clipping method,
we carefully searched clipping parameter $c$ and confirmed that
the best value is $c = 0.01$. We set $n_D$ to 1 for the both methods.

%

\subsection{Comparative methods}
For comparison, we implemented two models:
(i) a simple model in which the generator has one linear layer and four 3D deconvolutional layers
and the discriminator has five 3D convolutional layers,
and (ii) a Video GAN proposed by \cite{Vondrick16}.
We call the former {\it ``3D model''}.
In the generator of the 3D model, all the deconvolutional layers
have $4 \times 4 \times 4$ kernel and the stride of 2.
The number of channels in the initial deconvolutional layer is 512
and set to half when the layer goes deeper.
We also used ReLU and batch normalization layers.
The settings of the discriminator are exactly the same as
those of our model. In the settings of the video GAN,
we simply followed the settings in the original paper.

When we tried to train the 3D model and the video GAN model with the normal GAN loss,
we observed that the discriminator easily wins against the generator
and the training cannot proceed.
To avoid this, we added Gaussian noise ($\sigma = 0.2$) to all layers of discriminators.
In this case, all the scale parameters $\gamma$ after the Batch Normalization layer are not used.
Note that this noise addition is not used when we use the WGAN.

\subsection{Qualitative evaluation}
\begin{figure}[t]
\centering
\hfil\renewcommand{\arraystretch}{0.7}
$\begin{array}{cc}
\parbox[c][5mm][c]{0mm}{}
\!\!\!\!\text{\small Frame 1 \hspace{17mm} Frame 16} &
\!\!\!\!\text{\small Frame 1 \hspace{17mm} Frame 16} \\
\!\!\!\!\includegraphics[height=21mm]{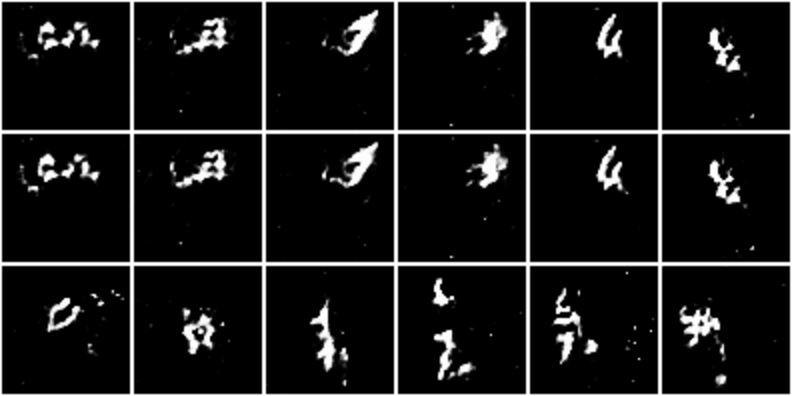} &
\!\!\!\!\includegraphics[height=21mm]{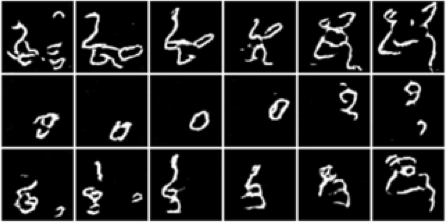} \\
\parbox[c][5mm][c]{0mm}{}
\!\!\!\!\text{\small (a) 3D model (GAN)} &
\!\!\!\!\text{\small (b) 3D model (WGAN w/ SVC)} \\
\!\!\!\!\includegraphics[height=21mm]{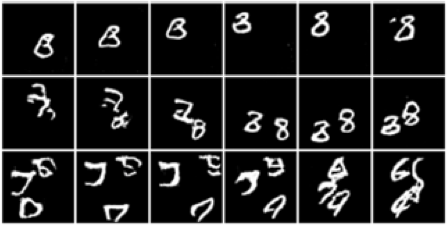} &
\!\!\!\!\includegraphics[height=21mm]{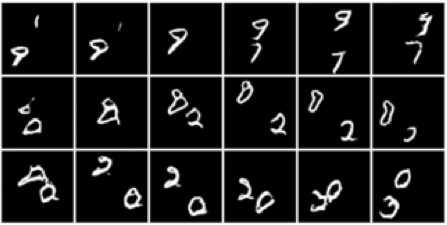} \\
\parbox[c][5mm][c]{0mm}{}
\!\!\!\!\text{\small (c) TGAN (SVC, $G_1(z_1^t)$)}&
\!\!\!\!\text{\small (d) TGAN (SVC, $G_1(z_0, z_1^t)$)} \\
\end{array}$
\caption{Generated videos with four different models:
(a) 3D model trained with the normal GAN,
(b) 3D model trained with the WGAN and the SVC,
(c) TGAN in which $G_1$ only uses $z_1$, and
(d) TGAN in which $G_1$ uses both $z_0$ and $z_1$.
Although these models generate 16 frames, for brevity
we extract six frames from them at even intervals.}
\label{fig:MovingMNIST}
\end{figure}
\begin{figure}[t]
\centering
\hfil\renewcommand{\arraystretch}{0.7}
$\begin{array}{cc}
\!\!\includegraphics[width=38mm]{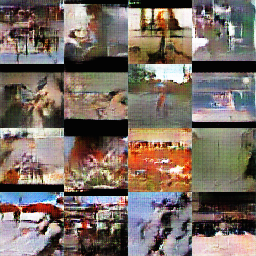} &
\!\!\includegraphics[width=38mm]{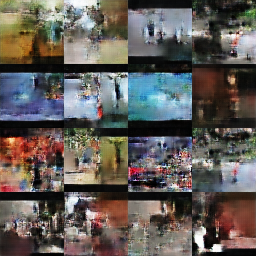} \\
\parbox[c][5mm][c]{0mm}{}
\!\!\text{\small (e) 3D model (Normal GAN)} &
\!\!\text{\small (f) 3D model (SVC)} \\
\!\!\includegraphics[width=38mm]{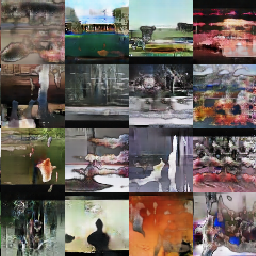} &
\!\!\includegraphics[width=38mm]{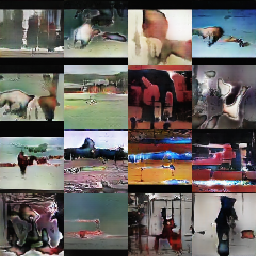} \\
\parbox[c][5mm][c]{0mm}{}
\!\!\text{\small (g) Video GAN (SVC)} &
\!\!\text{\small (h) TGAN (SVC)} \\
\end{array}$
\caption{A comparison between four models:
(e) 3D model trained with the normal GAN,
(f) 3D model trained with the WGAN and the SVC,
(g) Video GAN trained with the WGAN and the SVC,
and (h) TGAN trained with the WGAN and the SVC.
Only the first frame is shown.}
\label{fig:UCF101Images}
\end{figure}
We trained our proposed model on the above datasets and
visually confirmed the quality of the results. \Fig{MovingMNIST} shows
examples of generated videos by the generator trained on the moving
MNIST dataset.
It can be seen that the generated frames are quite different from those of the existing model proposed by Srivastava \etal \cite{Srivastava15}.
While the predicted frames by the existing model tend to
be blurry, our model is capable of producing consistent frames
in which each image is sharp, clear and easy to
discriminate two digits.
We also observed that although our method can generate the 
frames in which each digit continues to move in a straight 
line, its shape sometimes slightly changes by time.
Note that the existing models such as 
\cite{Srivastava15,Kalchbrenner16} 
seem to generate frames in which each digit does not change, 
however, these methods can not be directly compared with our method 
because the qualitative results the authors have shown are for ``video 
prediction'' that predicts future frames from initial inputs, whereas our 
method generates them without such priors.

\Fig{MovingMNIST} also shows that as for the quality of the 
generated videos, the 3D model using the normal GAN is the 
worst compared with the other methods.
We considered that it is due to the high degree of freedom
in the model caused by three-dimensional convolution, and
explicitly dividing the spatio-temporal space could contribute
to the improvement of the quality.
We also confirmed that it is not the effect of selecting
the normal GAN; although the quality of samples generated by 
the 3D model with the SVC outperforms that of the 3D model 
with the normal GAN, it is still lower than our proposed model
(model (d) in \Fig{MovingMNIST}).
In order to illustrate the effectiveness of $z_0$ in $G_1$,
we further conducted the experiment with the TGAN in which
$G_1$ does not take $z_0$ as an argument (model (c)).
In this experiment, we observed that in the model (c) the problem of mode collapse tends to occur compared to our model.

We also compared the performance of our method with other existing methods
when using practical data sets such as UCF-101.
The qualitative experimental results are shown in \Fig{UCF101Images}.
We observed that the videos generated by the 3D model have the most artifacts
compared with other models.
The video GAN tends to avoid these artifacts because the background is relatively 
fixed in the UCF-101, however, the probability of generating unidentified 
videos is higher than that of the proposed model.
We inferred that this problem is mainly due to the weakness of the existing 
method is vulnerable to videos with background movement.

\begin{figure}[t]
\centering
\hfil\renewcommand{\arraystretch}{0.7}
$\begin{array}{cc}
\!\!\!\!\includegraphics[height=21mm]{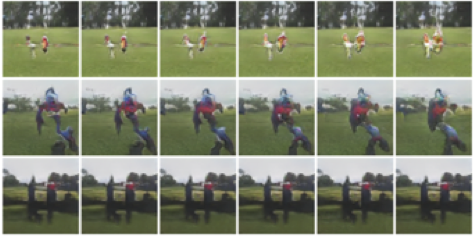} &
\!\!\!\!\includegraphics[height=21mm]{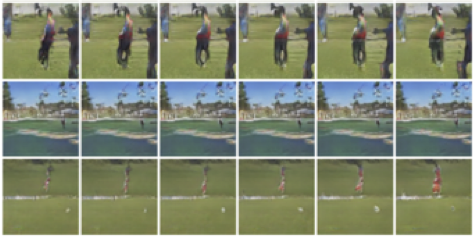} \\
\end{array}$
\caption{Example of videos generated by the TGAN with WGAN and SVC. The golf scene dataset was used.}
\label{fig:Golf}
\end{figure}
Finally, in order to indicate that the quality of our model is comparable
with that of the video GAN (these results can be seen in their project page),
we conducted the experiment with the golf scene 
dataset. As we described before, it is considered that this dataset, in which the background is always fixed, is advantageous for the video GAN that exploits
this assumption. Even under such unfavorable conditions, the quality of the videos generated by our model is almost the same as the existing method;
both create a figure that seems likes a person's shadow, and it changes with time.

\subsubsection{Applications}
\label{sec:ThreeApplications}
\begin{figure}[t]
\centering
\includegraphics[width=1.0\linewidth]{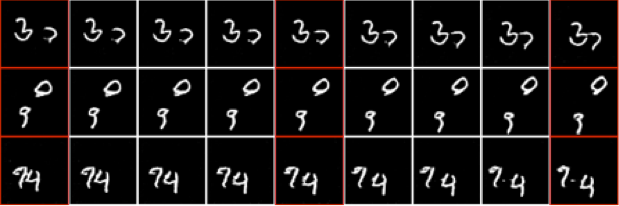}
\caption{Examples of frame interpolation with our method. The 
red columns represent the adjacent frames generated by the 
temporal generator.
The remaining columns show the intermediate frames.}
\label{fig:Interpolation}
\end{figure}

\begin{figure}[t]
\centering
\hfil\renewcommand{\arraystretch}{0.9}
$\begin{array}{lcc}
\!\!\!\rotatebox{90}{\hspace*{1mm} \text{IceDancing}} &
\!\!\!\!\includegraphics[width=0.47\linewidth]{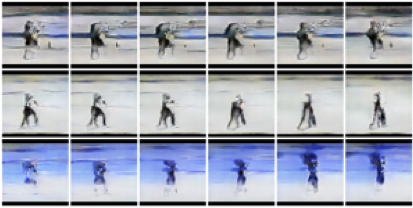} &
\!\!\!\!\includegraphics[width=0.47\linewidth]{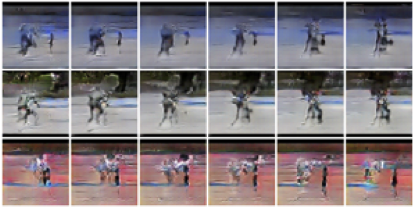} \\
\!\!\!\rotatebox{90}{\hspace*{0mm} \text{BaseballPitch}} &
\!\!\!\!\includegraphics[width=0.47\linewidth]{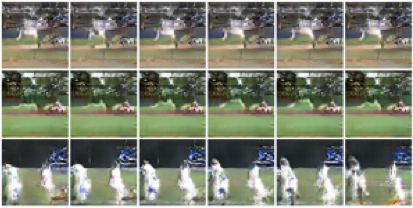} &
\!\!\!\!\includegraphics[width=0.47\linewidth]{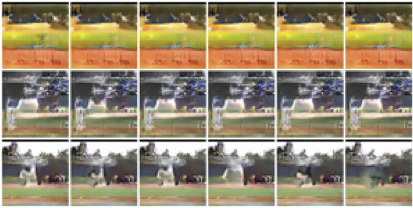} \\
\end{array}$
\caption{Generated videos by the conditional TGAN.
The leftmost column shows the category in UCF-101 dataset,
and the second and third columns show the generated samples
given the category.}
\label{fig:ConditionalTGAN}
\end{figure}

We performed the following experiments to illustrate the effectiveness of the applications described in \Sec{Applications}.

To show our model can be applied to frame interpolation,
we generated intermediate frames
by interpolating two adjacent latent variables of the image space.
These results are shown in \Fig{Interpolation}.
It can be seen that the frame is not generated by a simple interpolation 
algorithm like dissolve, but semantically interpolating the adjacent frames.

We also experimentally confirmed that the proposed model is also extensible to
the conditional GAN. These results are shown in \Fig{ConditionalTGAN}.
We observed that the quality of the video generated by the conditional TGAN
is significantly higher than that of the unsupervised ones.
It is considered that adding semantic information of labels to the model
contributed to the improvement of quality.

\subsection{Quantitative evaluation}
\begin{table}[t]
\centering
{\renewcommand{\arraystretch}{1.0}
\begin{tabular}{llll}
Model A & Model B & GAM score & Winner \\ \hline
TGAN & 3D model (GAN) & $1.70$ & TGAN \\
TGAN & 3D model (SVC) & $1.27$ & TGAN \\
TGAN & TGAN ($G_1(z_1^t)$) & $1.03$ & TGAN \\ \hline
\end{tabular}
}
\vspace{3pt}
\caption{GAM scores for models of moving MNIST.
``TGAN'' denotes the model trained with the WGAN and the SVC.
In ``TGAN ($G_1(z_1^t)$)'', $G_1$ has $z_1$ only (the SVC was used for training).
``3D model (GAN)'' and ``3D model (SVC)'' were trained with the normal GAN and the SVC, respectively.
}
\label{table:GAMScore}
\vspace{-10pt}
\end{table}
We performed the quantitative experiment to confirm the effectiveness
of our method. As indicators of the quantitative evaluation,
we adopted a {\it Generative Adversarial Metric (GAM)} \cite{Im16}
that compares adversarial models against each other,
and an {\it inception score} \cite{Salimans16} that
has been commonly used to measure the quality of the generator.

For the comparison of two generative models, we used GAM scores
in the moving MNIST dataset. Unlike the normal GAN in which the discriminator
uses the binary cross entropy loss, the discriminator of the WGAN is
learned to keep the fake samples and the real samples away, and we cannot choose zero as a threshold for discriminating real and fake samples.
Therefore, we first generate a sufficient number of fake samples,
and set a score that can classify fake and real samples well as the threshold.

\Table{GAMScore} shows the results. In the GAM, a score higher than one
means that the model A generates better fake samples that can fool the 
discriminator in the model B. It can be seen that our model can
generate better samples that can deceive other existing methods.
It can be seen that the TGAN beats the 3D models easily,
but wins against the TGAN in which $G_1$ has $z_1^t$ only.
These results are the same as the results obtained by
the aforementioned qualitative evaluation.

In order to compute the inception score, a dataset having label information
and a good classifier for identifying the label are required.
Thus, we used the UCF-101 dataset that has 101 action categories,
and a pre-trained model of C3D \cite{Tran15}, which was trained on Sports-1M 
dataset \cite{Karpathy14} and fine-tuned for the UCF-101,
was employed as a classifier.
We also calculated the inception scores by sampling 10,000 times from the latent 
random variable, and derived rough standard deviation by repeating this 
procedure four times.
To compute the inception score when using the conditional TGAN, we added the prior distribution for the category to the generator,
and transformed the conditional generator into the generator representing the model distribution. We also computed the inception score when using a real dataset to see an upper bound.

\begin{table}[t]
\centering
{\renewcommand{\arraystretch}{1.0}
\begin{tabular}{ll}
Method & Inception score \\ \hline
3D model (Weight clipping) & $4.32 \pm .01$ \\
3D model (SVC) & $4.78 \pm .02$ \\
Video GAN \cite{Vondrick16} (Normal GAN) & $8.18 \pm .05$ \\
Video GAN (SVC) & $8.31 \pm .09$ \\ \hline
TGAN (Normal GAN) & $9.18 \pm .11$ \\
TGAN (Weight clipping) & $11.77 \pm .11$ \\
TGAN (SVC) & $\boldsymbol{11.85 \pm .07}$ \\
Conditional TGAN (SVC) & $\boldsymbol{15.83 \pm .18}$ \\ \hline
UCF-101 dataset & $34.49 \pm .03$ \\ \hline
\end{tabular}
}
\vspace{3pt}
\caption{Inception scores for models of UCF-101.}
\label{table:InceptionScore}
\vspace{-10pt}
\end{table}
\Table{InceptionScore} shows quantitative results.
It can be seen that in the 3D model, the quality of the generated videos is 
worse than the video GAN and our proposed model. Although we observed that
using the SVC slightly improves the inception score, its value is a little and 
still lower than that of the video GAN.
We also confirmed that the SVC is effective in the case of the video GAN, 
however, its value is lower than our models.
On the other hand, our models achieve the best scores compared with other 
existing methods. In addition to the video GAN, the TGAN using the SVC slightly 
outperformed the TGAN using the conventional weight clipping method.
Although the quality of the SVC is almost indistinguishable
compared with existing methods,
we had to carefully change the value of $c$ to achieve such quality.
We believe that our clipping method is not a tool for dramatically
improving the quality of the generator,
but a convenient method to reduce the trouble of adjusting hyper parameters
and significantly stabilize the training of the models.

\section{Summary}
We proposed a generative model that learns
semantic representation of videos and can generate image sequences.
We formulated the generating process of videos as a series of
(i) a function that generates a set of latent variables,
and (ii) a function that converts them into an image sequence.
Using this representation, our model can generate videos with better quality
and naturally achieves frame interpolation.
We also proposed a novel parameter clipping method, Singular Value Clipping (SVC),
that stabilizes the training of WGAN.

\ificcvfinal
\paragraph{Acknowledgements}
We would like to thank Brian Vogel, Jethro Tan, Tommi Kerola, and Zornitsa Kostadinova
for helpful discussions.
\fi

{\small
\bibliographystyle{ieee}
\bibliography{library}
}

\end{document}